\documentclass[11pt]{article}
\usepackage{coling2020}
\usepackage{times}
\usepackage{url}
\usepackage{latexsym}
\usepackage{enumitem}
\usepackage{hyperref}
\usepackage{makecell}
\usepackage{graphicx}
\usepackage{etoolbox}
\usepackage[T1]{fontenc}

\title{Data Processing and Annotation Schemes for FinCausal Shared Task}

\author{Yseop Lab \\
  Dominique Mariko,
  Estelle Labidurie,\\
  Yagmur Ozturk,
  Hanna Abi-Akl, \\
  Hugues de Mazancourt \\
  {\tt fin.causal.task@gmail.com} \\}

\date{01/12/2019}

\begin{document}

\maketitle

\begin{abstract}
This document explains the annotation schemes used to label the data for the FinCausal Shared Task \cite{Mariko:20}. This task is associated to the Joint Workshop on Financial Narrative Processing and MultiLing Financial Summarisation (FNP-FNS 2020), to be held at The 28th International Conference on Computational Linguistics (COLING'2020), on December 12, 2020. \\

\end{abstract}

\section{Introduction}
\label{description}
Causality detection is a well known topic in the NLP and linguistic communities and has many applications in information retrieval. It has been studied extensively in a wide range of disciplines and domains knowledge, yet experts often disagree on the characterisation of a sufficient causal link, as causality can be expressed using many different syntactic patterns as well as contrasted semantic representations.
This shared task proposes data to experiment causality detection, and focuses on determining causality associated to an event. An event is defined as the arising or emergence of a new object or context in regard of a previous situation. So the task will emphasise the detection of causality associated with financial or economic analysis and resulting in a quantified output.

\section{Data}
\subsection{Data Processing}
The data are extracted from a corpus of 2019 financial news crawled by Qwam. The original raw corpus is an ensemble of HTML pages corresponding to daily information retrieval from financial news feed. These news mostly inform on the 2019 financial landscape, but can also contain information related to politics, micro economics or other topic considered relevant for finance information.

\subsection{Data License}
Data are released under the CC0 License

\section{Annotation scheme}

\subsection{Definition of causality}

A causal relationship involves the statement of a \textbf{cause} and its \textbf{effect}, meaning that two events or actors are related to each other with one triggering the other. 
We focused our annotation on text sections\footnote{We are using the term \textit{text section} since it could be a phrase, a sentence as well as a paragraph in which the cause and the effect are split in different sentences. For instance "Selling and  marketing  expenses decreased to \$1,500,000 in 2010. This was primarily attributable to employee-related actions and lower travel costs." However, in order to have a reproducible annotation process, we reduced the context to a paragraph of maximum three sentences.} that state causal relationships involving a quantified fact, which was necessary to reduce the complexity of the task. The following table displays the terms we use in the context of the Shared Task.

\begin{table}[h]
\begin{center}
\resizebox{\textwidth}{!}{%
\begin{tabular}{|l|l|}

\hline
\multicolumn{2}{|c|}{FACT} \\
\hline
\hline
Empirical Fact & Past event, acknowledged  \\
Process & Concrete event in duration  \\
State of Affairs & In being situation (will become true or false) \\
Looking Forward Statement & Expectation (often a declaration made by CPY board member) \\
Hypothesis & Projection based on facts \\
\hline
\hline
\multicolumn{2}{|c|}{QUANTIFIED FACT (QFact)} \\ 
\hline
\hline
Explicit &  Has a direct connection to an explicit measure \\
Measurable & Measure is either a quantity or a number that can be precisely identified in a text section \\
Verifiable & Is a State of Affairs at least \\
\hline

\end{tabular}%
}
\end{center}
\caption{\label{font-table} Representation of events terminology}
\end{table}

In this scheme, an effect can only be a quantified fact. The cause can either be a fact or a quantified fact. The causality between these two elements can be implicit as well as explicitly stated with a triggering linguistic mark also called a connective. The place of these sub-strings in the text section can vary according to the connective used or simply according to the author's style.
\\\\

In order to delimit the process, the distance between a cause fact and an effect fact was restricted to \textbf{a 3-sentences distance}. In other words, we only annotated a causal relationship when there was a maximal gap of 1 untagged sentence between the two facts. \\
As a matter of example: \\
"<cause>\textit{Previous management sought to transform the company from a simple milk processor into a producer of value-added dairy products as it chased profits offshore}<cause>.<effect>\textit{Among Fonterra's biggest missteps was the 2015 purchase of an 18.8 per cent stake in Chinese infant formula manufacturer Beingmate Baby \& Child Food for \$NZ755 million, just as the China market became hyper-competitive and demand slowed}<effect>. \textit{Fonterra last month announced it would cut its Beingmate stake by selling shares after failing to find a buyer. Meanwhile, back home, Fonterra's share of the milk processing market dropped from 96 per cent in 2001 to 82 per cent currently, with consultants TDB Advisory expecting it to be about 75 per cent by 2021."}\\ In this example, \textit{“...the 2015 purchase of an 18.8 per cent stake in Chinese infant formula manufacturer Beingmate Baby \& Child Food for \$NZ755 million”} was annotated as effect because this effect is within a 2 sentences away distance from the cause sentence. On the other hand, \textit{“Fonterra's share of the milk processing market dropped from 96 per cent in 2001 to 82 per cent currently”} was not annotated because this effect is 4 sentences away from the cause.

\subsection{Connectives}

A connective can be a verb, a preposition, a conjunction, an element of punctuation, or anything else, which explicitly introduces a causal relationship. Among those, there is a specific type of connective that is not taken into account in this Shared Task called lexical causative~\cite{Levin:15} . A lexical causative is a causal relationship stated through connectives (generally predicates) which, from a semantic point of view, also bear the effect of the cause. For instance in "The company raised its provisions by 5\% in 2018.", \textit{raise} is a lexical causative that can be glossed as \textit{The company caused the provisions to rise  by 5\%}. We will not consider those as causal references, since the effects are \textit{implied} in the connectives' definition. \par
Causal relationships can be introduced by other types of connectives in the identified text section. It is often rendered with the use of polysemous connectives which main function is not to introduce a causal relationship. For example, in this sentence: \textit{"Zhao found himself 60 million yuan indebted after losing 9,000 BTC in a single day (February 10, 2014)"}, the main function of the connective \textit{after} is to express a temporal relation between the two clauses. But we also have a causal relationship between them, since one triggers the other. \par
In the tagging process, the connectives involved in the causal relationship \textbf{were not annotated as part of the facts}. For example : <effect>\textit{Titan has acquired all of Core Gold's secured debt for \$US2.5 million}<effect> in order to <cause> \textit{ensure the long-term success of its assets.}<cause>. 
Two exceptions to this scheme are inserting the connectives in the cause or effect:

\begin{itemize}
    \item{the connective is inserted in the annotated fact, i.e. : <cause>\textit{On August 30, 2013, ST Yushun, \textbf{in order to} strengthen its competitive strength}<cause>, <effect>\textit{acquired a 100\% stake in ATV Technologies for 154 billion yuan}<effect>.}
    \item{the connective is at the beginning of a spanned linguistic unit, i.e. <cause>\textit{\textbf{As} Nvidia Com (NVDA) Stock Value Declined}<cause>, <effect>\textit{Shareholder Assetmark Has Cut Stake by \$15.67}<effect> } (see \underline{Third rule} in \ref{priority} Priority Rules) 
\end{itemize}

\subsection{Complex causal relationships}

In a text section, complex causal relationships can be rendered with conjoined relationships. A conjoined causal relationship can be one cause related to several effects, or one effect caused by several causes. This is often the case when the facts are not repeated and a conjunction is used as a link for the different effects or causes. This phenomenon can be also found in an implicit causal relationship and/or at sentence level. Here is an instance of a conjoined effect related to two causes: <cause>\textit{India's government slashed corporate taxes on Friday}<cause>\textit{,} <effect>\textit{giving a surprise \$20.5 billion break}<effect> <cause>\textit{aimed at reviving private investment and lifting growth from a six-year low that has caused job losses and fueled discontent in the countryside}<cause>. In the tagging process, they were first annotated as separate facts and then grouped according to priority rules if any applied (see \underline{Fourth rule} and \underline{Fifth rule} in \ref{priority} Priority Rules)

\subsection{Priority rules}
\label{priority}
The priority rules allow the annotation process of causal relationships to be more accurate and harmonious \footnote{As it might sometimes be difficult to distinguish sentences form strings in the designed blocks, we added the index of the offset we used in the Task 2 dataset to segment text blocks into sentences and apply priority rules.}.\\

\underline{First rule.} If a sentence contained \textbf{only one fact} (cause or effect), we \textbf{tagged the entire sentence} (even if it contains some noise or a connective). For instance : <cause> \textit{Hurricane Irma was the most powerful storm ever recorded in the Atlantic and one of the most powerful to hit land, Bonasia said.}<cause><effect>\textit{It cause \$50 billion in damages.}<effect>\\\

\underline{Second rule.} The \textbf{annotation of sentence-to-sentence causal relationships is prioritized}. When the annotator had the choice between linking two full sentences together or subdividing a sentence, he chose the sentence-to-sentence annotation. To illustrate this point, let's look at the text section:\\
"\textit{Finally, Seizert Capital Partners LLC increased its holdings in shares of BlackRock Enhanced Global Dividend Trust by 17.2\% during the second quarter. Seizert Capital Partners LLC now owns 138,020 shares of the financial services provider's stock valued at \$1,481,000 after acquiring an additional 20,223 shares in the last quarter.} \\
In this text section, there are two causal relationships. The first one links “\textit{Seizert Capital Partners LLC increased its holdings in shares of BlackRock Enhanced Global Dividend Trust by 17.2\% during the second quarter}” and “\textit{Seizert Capital Partners LLC now owns 138,020 shares of the financial services provider's stock valued at \$1,481,000}”. \\ Since the two facts are located into different sentences, we would have to annotate the full sentences each time (rule 1). 
\\The second causal relationship links “\textit{Seizert Capital Partners LLC now owns 138,020 shares of the financial services provider's stock valued at \$1,481,000}” and "\textit{acquiring an additional 20,223 shares in the last quarter}". Here, a sentence is subdivided. \par
Considering the priority of sentence-to-sentence annotation, the final annotation of this text section was: "<cause>\textit{Finally, Seizert Capital Partners LLC increased its holdings in shares of BlackRock Enhanced Global Dividend Trust by 17.2\% during the second quarter}<cause>. <effect>\textit{Seizert Capital Partners LLC now owns 138,020 shares of the financial services provider's stock valued at \$1,481,000 after acquiring an additional 20,223 shares in the last quarter.}<effect>"\par
This rule also highlights that \textbf{two different annotations cannot overlap}. We choose not to annotate the following causal chain: "<cause>\textit{acquiring an additional 20,223 shares in the last quarter}"<cause> and <effect>\textit{Seizert Capital Partners LLC now owns 138,020 shares of the financial services provider's stock valued at \$1,481,000 after acquiring an additional 20,223 shares in the last quarter.}<effect>, because these two texts segments reconcile into a higher level <effect> in our annotation scheme \\\

\underline{Third rule.} \textbf{When a causal chain is located inside a single sentence, in order to facilitate the extraction process, we chose to span the causal units as much as possible}, i.e. considering the following exact causal units:\\
"This week's bad news comes from Rothbury, Michigan, where<cause>\textit{Barber Steel Foundry will close at the end of the year} <cause>, <effect>\textit{leaving 61 people unemployed}<effect>", \\The spans were extended in order to cover the entire sentence. Only the connector, when located in between the cause and the effect, was left out of the extraction. As a result, the final causal chain is: \\ <cause>\textit{This week's bad news comes from Rothbury, Michigan, where Barber Steel Foundry will close at the end of the year} <cause>, <effect>\textit{leaving 61 people unemployed}<effect>. The spanning extension facilitate the consistency of the annotation process.\\\

\underline{Fourth rule.} If \textbf{two facts of the same type were located in the same sentence and were related to the same effect or cause}, then \textbf{we annotated these two facts as one unit}. For instance, in the text section: \\ "\textit{Thomas Cook's demise leaves its German operations hanging. More than 140,000 German holidaymakers have been impacted and tens of thousands of future travel bookings may not be honored.}", the cause fact is “\textit{Thomas Cook's demise}”. \\Since it was the only fact in the sentence, we annotated the full sentence as the cause (see priority rule number 1). The cause fact has two consequences: “\textit{More than 140,000 German holidaymakers have been impacted}” and “\textit{tens of thousands of future travel bookings may not be honored}”. \\ Since both effect facts are in the same sentence and related to the same cause, we annotated the text section as follow : <cause>\textit{Thomas Cook's demise leaves its German operations hanging.}<cause><effect>\textit{More than 140,000 German holidaymakers have been impacted and tens of thousands of future travel bookings may not be honored}<effect>.\par
This rule was also applied to the annotation of cause.s and effect.s inside a sentence. For instance :\\ "<effect>\textit{Our total revenue decreased to \$31 million}<effect> <cause>\textit{due to decrease in orders from approximately \$91,000 to \$82,000, and a decrease in total buyers, which includes both new and repeat buyers from approximately 62,000 to 56,000.}<cause>". The two causes were put together since they are related to the same effect.\par
This rule was only used in the two cases presented above. When more than two sentences were involved it was not taken into account. For example : "<cause>\textit{Let's say Shirley reduced her assets of \$165,000 through a gift of \$10,000 and pre-paying her funeral expenses for \$15,000.}<cause> <effect1>\textit{Her DAC would reduce from \$55 a day to \$43 a day (a saving of just over \$4,300 a year).}<effect1><effect2>\textit{Her equivalent lump sum would reduce by almost \$88,000!}<effect2>". Consequently, the same text section may appear twice in the release dataset.
\\\\\

\underline{Fifth rule.} The annotation of \textbf{causal chains} inside a sentence. A segment of text that is a cause can also be the effect of another cause. For instance, the sentence \\"\textit{BHP emitted 14.7m tonnes of carbon dioxide equivalent emissions in its 2019 fiscal year, down from 16.5m tonnes the previous year due to greater use of renewable energy in Chile.}" contains three facts: \\ "\textit{greater use of renewable energy in Chile} is the cause of \textit{down from 16.5m tonnes the previous year} which is also the cause of \textit{BHP emitted 14.7m tonnes of carbon dioxide equivalent emissions in its 2019 fiscal year}.\par
In that case, we \textbf{isolated the rightmost fact and tagged it according to its nature}. All \textbf{the remaining facts were gathered as one unit and annotated with the remaining tag}. In the above example this rule eventually provides this final annotation : "<effect>\textit{BHP emitted 14.7m tonnes of carbon dioxide equivalent emissions in its 2019 fiscal year, down from 16.5m tonnes the previous year}<effect> <cause>\textit{greater use of renewable energy in Chile}<cause>"

\subsection{Other annotation levels}
The cause or the effect can sometimes be found as pronouns, relative pronouns included. In that case, the reference (the antecedent) of the pronoun, is the extracted element. For instance, in the text: \\"The tax revenues decreased by 0.3\%, which was caused by fiscal decentralization reform." \textit{The tax revenue decreased by 0.3\%} corresponds to the effect and \textit{fiscal decentralization reform} is the cause. \\In some cases, the pronoun can be added to the opposite fact where the antecedent is.
\\\\
The role of a clause in a causal sentence can be ambiguous to identify. For example, it can be precarious to tell whether the clause corresponds to the cause, the means or the goal. If so, the sequence was annotated as the cause.\par
The ambiguity can also exist between two facts - which is the cause? which is the effect? In that case, when there was only one Qfact, the latter was annotated as the effect. When both facts were Qfacts, the annotation order was left to the annotator's appreciation. The annotator was encouraged to use reformulation in order to decide which fact was the cause and which fact was the effect.
\\\\
If the cause is in the middle of the effect or vice versa, the sentence is not annotated because of the conflict process. Here is an example: "The take-home pay after necessary deductions is S\$4,137." where \textit{after necessary deductions} is a cause inserted in the effect.
\\\\
We decided not to annotated causal relationships with structures identical to a calculation structure. For instance, in the text section "\textit{Google has 100K+ people and \$136B in revenue (2018), earning over \$1.3M per person.}", we considered that, since the quantified fact \textit{earning over \$1.3M per person.} is the result of a calculation that can be recomputed from data available in the cause fact, it triggers no new information. Consequently, it was not considered a causal chain and was not annotated.
\\\\
Finally, dates are also to be included in the fact annotated if it is related to it and is placed next to it in the sentence.

\section{Conclusion}
The final annotated data are to be presented during the  Joint Workshop on Financial NarrativeProcessing and MultiLing Financial Summarisation (FNP-FNS 2020), at COLING online, December 12 2020, along the associated FinCausal Shared Task and results.

\end{document}